%% file: ICRA2022.tex
\documentclass[letterpaper, 10 pt, conference]{ieeeconf}
\usepackage{times}
\usepackage[pdftex]{graphicx}
\usepackage{subfigure}
\usepackage{amsmath,amssymb,amsopn,amstext,amsfonts}
\usepackage{cancel}
\usepackage[space]{cite}
\usepackage{pdfsync}
\usepackage{balance}
\usepackage{color}
\usepackage{mathtools}
\usepackage{multirow}
\usepackage{makecell}
\usepackage{algorithm}  
\usepackage{algorithmicx}  
\usepackage{algpseudocode} 
\usepackage{bm}

\usepackage{diagbox}
\usepackage{float}
\usepackage{epstopdf}
\usepackage{pifont}
\usepackage{multirow}
\usepackage{url}
\usepackage{tabularx}
\usepackage[linkcolor=black,citecolor=black,urlcolor=black,colorlinks=true]{hyperref}
\usepackage{verbatim} 
\usepackage[skip=3pt, font={small}]{caption}

\newcommand{\df}[1]{\mathrm{d}{#1}}

\newcommand{\norm}[1]{\Vert{#1}\Vert}

\graphicspath{{../figure/}}
\DeclareGraphicsExtensions{.png,.jpg,.eps,.pdf}
\IEEEoverridecommandlockouts
\title{\LARGE \bf GS-Planner: A Gaussian-Splatting-based Planning Framework \\   for Active High-Fidelity Reconstruction}
\author{Rui Jin$^*$, Yuman Gao$^*$, Yingjian Wang, Haojian Lu, and Fei Gao
	\thanks{ $^*$\textbf{Equal contribution.}
	All authors are with the Institute of Cyber-Systems and Control, College of Control Science and Engineering, Zhejiang University, Hangzhou 310027, China, and also with the Huzhou Institute, Zhejiang University, Huzhou 313000, China. {\tt\small \{fgaoaa\}@zju.edu.cn}.
	}
}

\makeatletter
\let\@oldmaketitle\@maketitle
\renewcommand{\@maketitle}{\@oldmaketitle
	\vspace{0.3cm}
	\centering
	\setcounter{figure}{0}
	\begin{minipage}{1.0\linewidth}
		\includegraphics[width=1.0\textwidth]{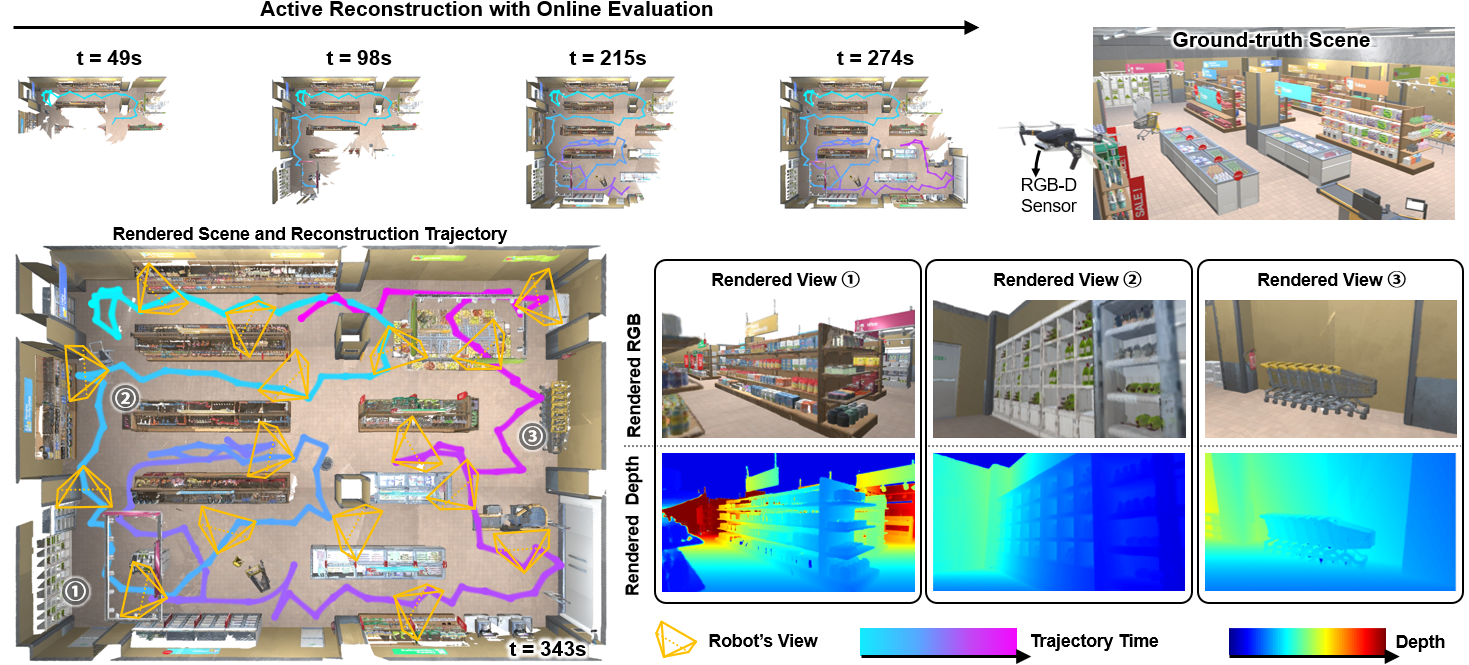}
		\vspace{-0.2cm}
		\captionof{figure}{\label{fig:simu_res}The whole active reconstruction process in a simulated supermarket scene. We deployed our active high-fidelity reconstruction system on a simulated quadrotor with an RGB-D sensor. The colored curves illustrate the executed trajectories of the drones. We demonstrate the reconstruction result including the whole rendered scene and details rendered at three views.}
	\end{minipage}
	\vspace{-0.0cm}
}

\makeatother

\begin{document}

\maketitle
\thispagestyle{empty}
\pagestyle{empty}


\begin{abstract}
Active reconstruction technique enables robots to autonomously collect scene data for full coverage, relieving users from tedious and time-consuming data capturing process.
However, designed based on unsuitable scene representations, existing methods show unrealistic reconstruction results or the inability of online quality evaluation.
Due to the recent advancements in explicit radiance field technology, online active high-fidelity reconstruction has become achievable.
In this paper, we propose GS-Planner, a planning framework for active high-fidelity reconstruction using 3D Gaussian Splatting.
With improvement on 3DGS to recognize unobserved regions, we evaluate the reconstruction quality and completeness of 3DGS map online to guide the robot. 
Then we design a sampling-based active reconstruction strategy to explore the unobserved areas and improve the reconstruction geometric and textural quality. 
To establish a complete robot active reconstruction system, we choose quadrotor as the robotic platform for its high agility.
Then we devise a safety constraint with 3DGS to generate executable trajectories for quadrotor navigation in the 3DGS map.
To validate the effectiveness of our method, we conduct extensive experiments and ablation studies in highly realistic simulation scenes.

\end{abstract}

\section{Introduction}
\label{sec:introduction}

Active high-fidelity 3D reconstruction involves robots creating an accurate, detailed, and realistic digital representation of an object or scene completely, efficiently, and safely. 
Maintaining intricate visual fidelity, this technique demonstrates significant practical value in scene inspection, virtual game development, and cultural heritage preservation.

The choice of an appropriate scene representation is the cornerstone of an active high-fidelity 3D reconstruction robotic system, with the following key requirements:


\begin{figure*}[t]
	\begin{center}
		\includegraphics[width=2.0\columnwidth]{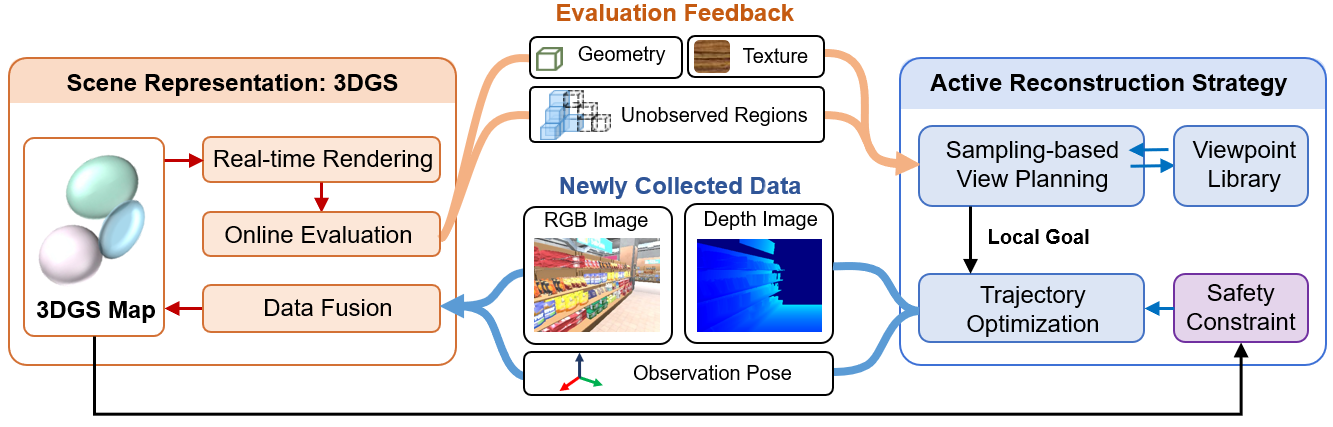}
	\end{center}
	\vspace{-0.0cm}
	\caption{
		\label{fig:system_overview} An overview of our active high-fidelity reconstruction system. With 3DGS as scene representation, the unobserved regions, as well as the geometric and textural information of the built map can be feedback in real time for online reconstruction quality and completeness online evaluation. The proposed active reconstruction strategy guides the robot to collect new scene information to build a complete high-fidelity 3DGS map.   }
	\vspace{-0.0cm}
\end{figure*}

\begin{itemize}
    \item \textbf{Precision and photorealism}: 
    High-fidelity reconstruction requires the scene representation to accurately represent geometric and textural information, which enables a more realistic portrayal of the scene.
    
    
    \item \textbf{Real-time fusion}: 
    New scene information is gathered step by step within the active reconstruction process. The scene representation should fuse the newly collected data in real time to guide the robot's reconstruction and provide information about occupied volumes for safe robot navigation.

    \item \textbf{Online evaluation}:
    To guide the robot in active reconstruction, the scene representation requires online evaluation of both reconstruction quality and completeness. The quality assessment should include both geometric and textural aspects. The completeness evaluation demands the representation to identify observed and unobserved portions of the scene. 

    
\end{itemize}

However, commonly used in traditional active reconstruction \cite{isler2016information, dang2019graph, zhou2021fuel, corah2019ral}, grid map can only describe coarse structures and lack color and texture information.  
Mesh and surfel cloud fusion and optimization pose challenges due to their inherent complexities.
Neural Radiance Field (NeRF) \cite{mildenhall2021nerf}, which recently emerged as a high-fidelity scene representation, often requires extensive training times and substantial resources for rendering, making online evaluation difficult.

3D Gaussian Splatting (3DGS) \cite{kerbl20233d}, which recently emerged as a transformative technique in the explicit radiance field, fully meets the above requirements with specific advantages as follows: 
(a) High visual quality and precise geometry: 
3DGS represents a scene with Gaussian blobs storing rich textural and explicit geometric information, ensuring high visual fidelity and precise geometry. More importantly, with learnable 3D Gaussians, 3DGS preserves properties of continuous volumetric radiance fields, which is essential for high-quality image synthesis. 
(b) Efficient fusion: 
Benefiting from explicit representation, 3DGS's frustum culling strategy and adaptive Gaussian densification make it efficient to incrementally fuse the new observed data, showing comparable quality and superior efficiency surpassing neural-based methods. 
(c) Fast rendering: 
3DGS's highly parallel "splatting" rasterization, along with the avoidance of the computational overhead associated with rendering in empty space, enables fast frame rates and high-quality rendering for online evaluation.



Due to 3DGS's appealing features, we propose a Gaussian-Splatting-based planning framework (GS-planner) to achieve active high-fidelity reconstruction with real-time quality and completeness evaluation to guide the robot's reconstruction.
Firstly, to evaluate the built 3DGS within the reconstruction process, we devise evaluation terms for both reconstruction completeness and quality. 
Existing 3DGS can only represent occupied space, making it difficult to evaluate the completeness. 
To efficiently identify unobserved portions of the scene, we integrate the unknown voxels into the splatting-based rendering process.
Secondly, we design a sampling-based active exploration strategy to guide the robot to explore the unobserved areas and improve the geometric and textural quality of the 3DGS map. 
Thirdly, to form a complete robotic active reconstruction system, we select quadrotor as the robotic platform for its high agility.
Leveraging the differentiable nature and explicit representation properties of 3DGS, we devise a differentiable obstacle-avoidance cost with the 3DGS map. Furthermore, we form an autonomous navigation framework capable of generating collision-free and dynamic-feasible trajectories for quadrotors. 
Overall, based on the state-of-the-art dense 3DGS SLAM system SplaTam \cite{keetha2023splatam}, we propose GS-Planner, a planning framework for active high-fidelity reconstruction with 3DGS as scene representation. 
In summary, the following are the contributions:
\begin{enumerate}
    \item We propose the first active 3D reconstruction system using 3DGS with online evaluation.
    \item We design evaluation metrics for reconstruction completeness and quality, applying them in a sampling-based autonomous reconstruction framework.
    \item We devise a safety constraint with 3DGS and form a trajecotry optimization framework in the 3DGS map.
    \item We conduct extensive simulation experiments to validate the effectiveness of the proposed system.
\end{enumerate}

\begin{figure*}[t]
    \begin{center}
        \includegraphics[width=2.0\columnwidth]{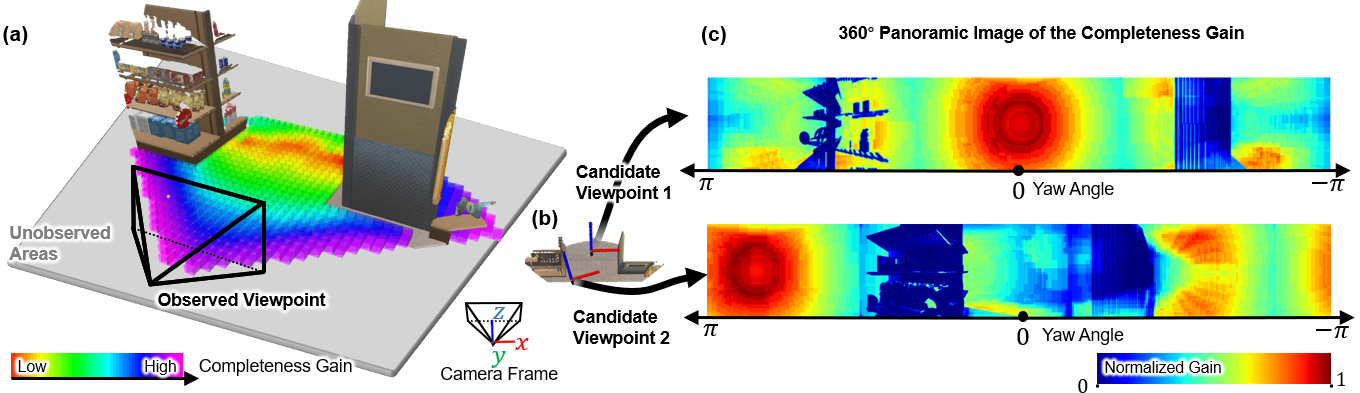}
    \end{center}
    \caption{An illustration of the completeness evaluation. (a). A partially reconstructed scene. Scene information has been collected only at the observed viewpoint. The colored grid illustrates the completeness gain from 360-degree summation at different positions at a height of $z=1m$. (b). The location of two candidate viewpoints. The z-axis direction is aligned with the camera's optical axis. (c). 360-degree panoramic image of the completeness gain of the candidate viewpoint 1 and 2. The generation of 360-degree gain facilitates the subsequent determination of the optimal viewpoint yaw direction.
}
    \label{fig:completeness}
    \vspace{-0.0cm}
\end{figure*}

\section{Related works}
\label{sec:related_works}

\subsection{High-fidelity Reconstruction}
\label{High-fidelity Reconstruction}
To achieve high-fidelity reconstruction, several different scene representations have been employed, such as planes, meshes, and surfel clouds.
Recently, Neural Radiance Field (NeRF) \cite{mildenhall2021nerf} has gained prominence in this field due to its exceptional capability of photorealistic rendering, which can be divided into three main types: MLP-based methods, hybrid representation methods, and explicit methods.
MLP-based method \cite{sucar2021imap} offers scalable and memory-efficient map representations but faces challenges with catastrophic forgetting in larger scenes. Hybrid representation \cite{zhu2022nice, jiang2023h2} methods combine the advantages of implicit MLPs and structure features, significantly enhancing the scene scalability and precision. As for the explicit method proposed in \cite{fridovich2022plenoxels}, it stores map features in voxel directly, without any MLPs, enabling faster optimization. 

While NeRF excelled in photorealistic reconstruction \cite{ran2023neurar}, NeRF methods are computationally intensive \cite{takikawa2021neural, reiser2021kilonerf, muller2022instant}. NeRF often requires extensive training times and substantial resources for rendering, which contradicts the need to feed the model back into the active reconstruction decision loop in real time. Instead of representing maps with implicit features, 3DGS \cite{kerbl20233d} enables real-time rendering of novel views by its pure explicit representation and the novel differential point-based splatting method.
This technology has been applied in online dense SLAM with 3DGS as the scene representation and reconstructs the scene from RGB-D images \cite{yan2023gs, keetha2023splatam}.

\subsection{Active Reconstruction System}
\label{Formation Maintenance}
Active reconstruction systems put data acquisition in the decision loop, using the results for evaluation, and then guiding the robot for further data acquisition. Based on the representations of 3D models, these approaches can be categorized into voxel-based methods \cite{isler2016information, dang2019graph, zhou2021fuel, corah2019ral}, surface-based approaches \cite{huang2018active, schmid2020efficient, gao2022meeting}, and neural-based approaches \cite{ran2023neurar}.

Voxel-based methods \cite{isler2016information, zhou2021fuel, corah2019ral} aim to reconstruct the commonly used grid map, which is an axis-aligned and compact spatial representation.
Surface-based approaches \cite{huang2018active, schmid2020efficient, gao2022meeting} model the environment with a set of surfaces. 
However, these methods only evaluate the reconstruction completeness but ignore color and texture information.
There are also active reconstruction methods based on implicit neural representations. NeurAR \cite{ran2023neurar} learns the neural uncertainty for view planning. However, limited by the high computation consumption of the implicit neural representation, NeurAR takes about 50-120 seconds for model optimization and uncertainty evaluation between view steps, leading to frequent and prolonged halts in robot operation.
3DGS, as a newly emerged method, is well-suited for serving as a scene representation for active high-fidelity reconstruction. However, there is currently no active reconstruction robot system designed based on its excellent characteristics.

\label{sec:formation similarity}
\section{System Overview}



Active high-fidelity reconstruction requires a robot to visit a series of viewpoints to collect scene information and build a realistic digital representation. 
As shown in Fig.~\ref{fig:system_overview}, the proposed active reconstruction system uses 3DGS as scene representation, and the robot can collect RGB-D images with the corresponding observation poses.
Leveraging the efficient fusion and real-time rendering advantages of 3DGS, we conduct an online evaluation for possible future viewpoints. 
Such online evaluation feedback guides the active view planning module (Sec.~\ref{sec:Active View Planning}) to generate a series of safe and high-information-gain viewpoints.
To navigate the robot to the selected viewpoints, we further propose an autonomous navigation framework (Sec.~\ref{sec: Trajectory Optimization}) with a safety constraint formulated with the 3DGS map.


\input{SEC4_method}
\input{SEC5_experiments}

\section{Conclusion and Future Work}
In this paper, we adopt the recently emerged 3DGS technique to achieve an active high-fidelity reconstruction system.
To online evaluate the reconstruction result as reconstruction strategy feedback, we respectively design completeness and quality evaluation methods with 3DGS.
Then we propose a sampling-based active view planning method to generate a series of optimal viewpoints.
For robot navigation in 3DGS map, we design a differentiable chance constraint to ensure safety, and form a quadrotor trajectory optimization framework.
For future work, we are going to deploy our system on real robotic platforms and try to reduce the GPU memory consumption of 3DGS and improve its efficiency.

\bibliography{ICRA2022}
\end{document}

%% file: SEC4_method.tex
\section{Active View Planning with 3DGS Map}
\label{sec:Active View Planning}
In this section, we first introduce the 3DGS representation (Sec.~\ref{subsec:3DGS Map Representation}). 
Then, a completeness evaluation method (Sec.~\ref{subsec:complete_eval}) and a quality evaluation method (Sec.~\ref{subsec:quality_eval}) are proposed to capture regions with poor coverage and quality respectively. 
In the following, we design a sampling-based active view planning algorithm to guide the robot to reconstruct unobserved regions and improve the quality of the built map (Sec.~\ref{subsec: view planning}).

\begin{figure*}[th]
    \begin{center}
        \includegraphics[width=2.0\columnwidth]{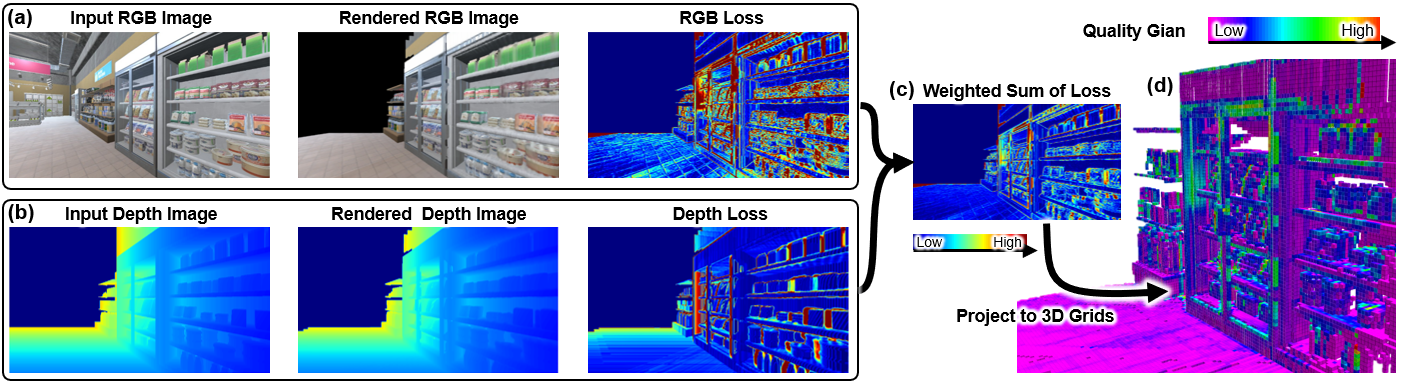}
    \end{center}
    \caption{An instance of the quality evaluation. (a). The generation of the RGB textural loss between the input RGB image and rendered RGB image. (b). The generation of the depth loss between the input depth image and rendered depth image. (c). The weighted sum of the RGB loss and depth loss. (d). We project the quality gain to the 3D grid in the world frame to store.}
    \label{fig:quality}
    \vspace{-0.0cm}
\end{figure*}

\subsection{3DGS Map Representation}
\label{subsec:3DGS Map Representation}
We use the existing method SplaTam SLAM \cite{keetha2023splatam} for 3DGS real-time reconstruction. The scene is represented as a set of isotropic 3DGS. Each 3D GS is defined by center position $\mu \in \mathbb{R} ^3$, radius $r \in \mathbb{R}$, RGB color $c \in \mathbb{R}^3$, and opcity $o \in \mathbb{R}$. The opacity function $\alpha$ of a point $x \in \mathbb{R} ^3$ computed from each 3DGS is described as:
\begin{equation}
\label{eq:opacity}
    \alpha\left(x, o\right)=o\exp{\left(-\frac{|x-\mu|^2}{2r^2}\right)}.
\end{equation}

In order to optimize the parameters of 3D Gaussians to represent the scene, we need to render them into images in a differentiable manner. The final rendered color can be formulated as the alpha-blending of N ordered points that overlap the pixel,
\begin{equation}
    C_{pix}=\sum_{i=1}^{N}{c_i\alpha_i\prod_{j=1}^{i-1}\left(1-\alpha_j\right)}.
\end{equation}
We render the depth in the same way
\begin{equation}
    D_{pix}=\sum_{i=1}^{N}{d_i\alpha_i\prod_{j=1}^{i-1}\left(1-\alpha_i\right)},
\end{equation}
where $d_i$ represents the depth of the $i$-th 3D Gaussian's center, which is equal to the z-coordinate of its center position in camera coordinate system.

\subsection{Completeness Evaluation}
\label{subsec:complete_eval}

To support full coverage of the scene, we introduce the completeness evaluation for candidate viewpoints. 
This evaluation requires to recognize unobserved space. However, the existing 3DGS only preserves information regarding the occupied volume.
To address this limitation, we maintain a voxel map to represent unobserved volume, and integrate it into the splatting-based rendering.
Then, we can efficiently calculate model-consistent pixel-level completeness gain within the 3DGS rendering process.

\begin{figure}[h]
    \begin{center}
        \includegraphics[width=0.9\columnwidth]{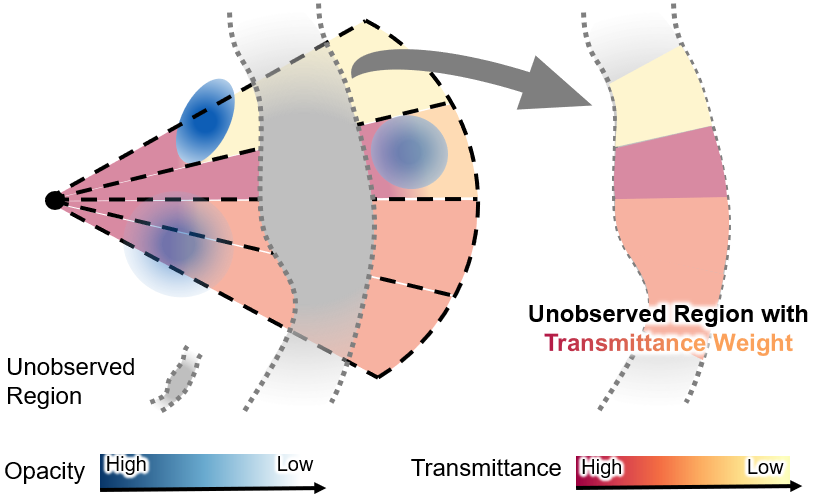}
    \end{center}
    \caption{A 2D illustration of the 3D completeness evaluation. Given a collection of 3D Gaussians and a candidate viewpoint, we can get unobserved regions within the splatting-based rendering. The unobserved regions are weighted by transmittance, which is equal to the accumulated Guassians' opacity along the ray.}
    \label{fig:unoberved}
    \vspace{-0.0cm}
\end{figure}

\begin{figure}[h]
    \begin{center}
        \includegraphics[width=0.9\columnwidth]{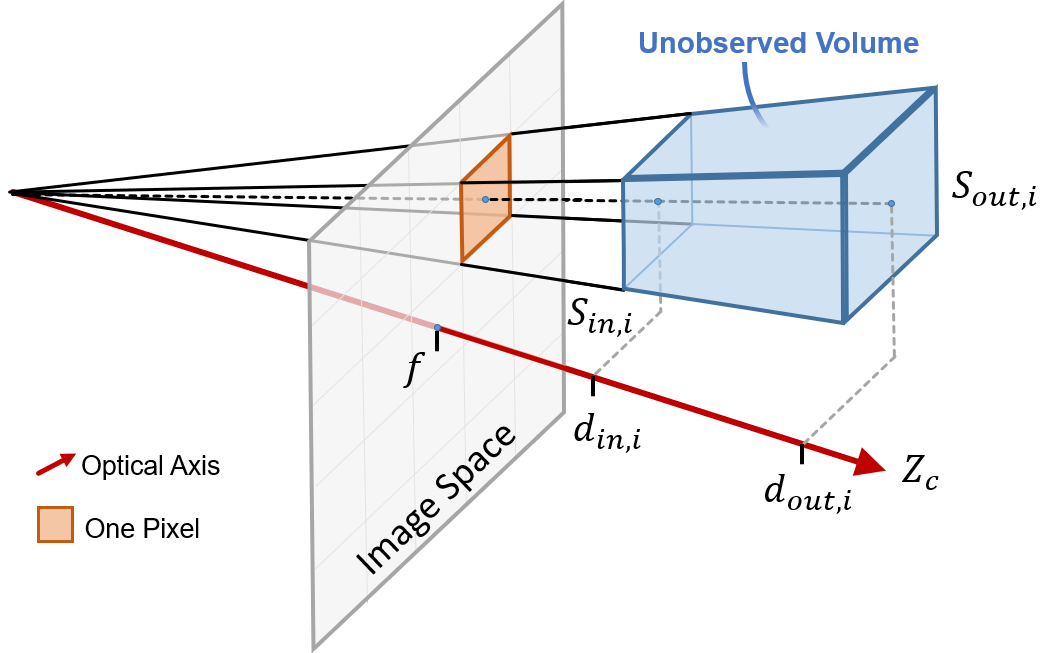}
    \end{center}
    \caption{A 3D illustration of the unobserved volume calculation.}
    \label{fig:frustum}
    \vspace{-0.0cm}
\end{figure}

To be specific, given a collection of 3D Gaussians and a candidate viewpoint, first all Gaussians will be sorted by their depth. With the sorted Gaussians, depth image can be efficiently rendered by alpha-compositing the splatted 2D projection of each Gaussian in order in pixel space.
In this rendering process, we can determine whether there exists an unobserved region between adjacent Gaussians utilizing the maintained unobserved voxel. 
And the volume of the unobserved region corresponding to each pixel can be approximately calculated by the basic frustum volume formula.
Furthermore, considering that Gaussians have different opacities, we evaluate the visibility of the unobserved volume by applying a transmittance weight, as shown in Fig.~\ref{fig:unoberved}.
Finally, we get the completeness information gain of each pixel as
\begin{equation}
    V_{pix}=\sum_{i=1}^{n}{V_i\prod_{j=1}^{m_i}\left(1-\alpha_j\right)},
\end{equation}
where $n$ is the number of unobserved volumes along the ray, $m_i$ is the number of the related 3D Gaussians before the $i$-th unobserved volume $V_i$, $\prod_{j=1}^{m_i}\left(1-\alpha_j\right)$ is the transmittance weight. 
For a certain unobserved volume $V_i$, as shown in Fig.~\ref{fig:frustum}, we approximate its volume as a frustum:
\begin{equation}
    V_i=\frac{1}{3}(S_{in,i}+\sqrt{S_{in,i} S_{out,i}}+S_{out,i})(d_{out,i}-d_{in,i}),
\end{equation} 
where $d_{in,i}$ and $d_{out,i}$ respectively represent the depths of the entry and exit of the $i$-th unobserved volume.
$S_{in,i}$ and $S_{out,i}$ represent the base areas of $V_i$, which are equal to the projected areas of the pixel at the entry plane of depth $d_{in,i}$ and the exit plane of depth $d_{out,i}$. $S=d^2/f^2$, where $f$ is the camera focal length.

Because we integrate the evaluation into the splatting-based rendering, this calculation process is parallelized and efficient.
To illustrate the completeness evaluation intuitively, we give an instance shown in Fig.~\ref{fig:completeness}, demonstrating the guidance of completeness evaluation in viewpoint selection.





\subsection{Quality Evaluation}
\label{subsec:quality_eval}
Quality evaluation aims to identify reconstructed regions with poor texture and geometry accuracy. This evaluation includes two steps: loss caching and loss reprojection.

Loss caching: Leveraging the real-time rendering of 3DGS, it is straightforward to compute the disparity between the reconstructed model and the actual scene. 
As shown in Fig.~\ref{fig:quality}, we project the loss $L$ from the image space to the world space, and cache the loss into occupied voxels.
Specifically, $L$ is a weighted sum of $L_1$ loss both on the depth and the color renders:
\begin{equation}
L = \mathrm{L}_1(D)+ \lambda_C \mathrm{L}_1(C),
\end{equation}
where $\lambda_C$ is the weight coefficient. 

Loss reprojection: Given a candidate viewpoint, we reproject the loss cached in the occupied voxels to the image space by conducting 360-degree ray-tracing. The loss indicates the quality information gain of both texture and geometry:

According to Sec.~\ref{subsec:complete_eval} and Sec.~\ref{subsec:quality_eval}, we finally obtain an overall 360-degree information gain of a given viewpoint by calculating the weighted sum of completeness and quality.
Then we use the sliding-window summation to find the optimal yaw angle of each viewpoint.

\subsection{View Planning with a View Library}
\label{subsec: view planning}
To enable a robot's full reconstruction of a scene in high quality, a series of reasonable viewpoints with position and yaw angle need to be generated for sequential navigation.
We design a sampling-based view planning method with a view library\cite{corah2019ral} to generate and cache viewpoints for evaluation. The whole view planning algorithm is listed as Alg.~\ref{alg1}.

To be specific, we first acquire nearby cached viewpoints $\mathbf{V}_{near}$ in the view library $\mathbf{VL}$, which stores unvisited viewpoints and their information gain (Line 1).
Their information gains are updated with new sensor data (Line 2-4).
We use the expansion part of RRT* to sample potential future viewpoints $\mathbf{V}_c$ (Line 5).
The sampled viewpoints that are too close to obstacles will be deleted.
And the optimal yaw angle of each viewpoint is determined by the above introduced sliding window method.  
$\mathbf{V}_{near}$ are added and connected to the expanded trees in the sampling process.
By real-time rendering at each viewpoint in $\mathbf{V}_c$ via 3DGS, we calculate its information gain efficiently (Line 6).
The viewpoints whose gain below threshold $g_{lb}$ will be removed (Line 8-12).
And high-information-gain viewpoints that are novel enough from others in $\mathbf{VL}$ will be cached (Line 13-15).
The node on the best branch will be selected as the next local goal (Line 19).
Moreover, if there are no valid nearby candidates, the local goal will be selected from $\mathbf{VL}$ (Line 21). 
When the $\mathbf{VL}$ becomes empty, the reconstruction process is accomplished.

\begin{algorithm}[t]
\caption{Active View Planning with a View Library}
\label{alg1}
\begin{algorithmic}[1]
	\Require current pose $\mathbf{p}$, view library $\mathbf{VL}$;


    \State $\mathbf{V}_{near}, \mathbf{G}_{near} \leftarrow$ \text{subset of $\mathbf{VL}$ nearby current pose $\mathbf{p}$};

    \For{$v_i \in \mathbf{V}_{near}, g_i \in \mathbf{G}_{near}$}
    \State $g_i$ = \textbf{UpdateGain}($v_i$);
    \EndFor

    \State $\mathbf{V}_c \leftarrow$ \textbf{RRTSample}($\mathbf{p}, \mathbf{V}_{near}$);
    \State $\mathbf{G}_c \leftarrow$ \textbf{Evaluation}($\mathbf{V}_c$);

    \For{$v_i \in \mathbf{V}_{c}, g_i \in \mathbf{G}_{c}$}
    \If{$g_i < g_{lb}$}
    \State $\mathbf{V}_c \leftarrow \mathbf{V}_c \setminus v_i$;
    \State $\mathbf{VL} \leftarrow \mathbf{VL} \setminus \{v_i,g_i\}$;
    \State \textbf{continue};
    \EndIf
    
    \If {\textbf{Overlap($v_i, \mathbf{V}_{near}$)} $< \epsilon_{ol}$ \textbf{and} $g_i > g_{thres}$} 
    \State $\mathbf{VL} = \mathbf{VL} \cup \{v_i,g_i\}$;
    \EndIf
    \EndFor

    \State Result local goal: $\mathbf{p}_{goal}$;

    \If {$\mathbf{V}_c != \emptyset$ }
    \State $\mathbf{p}_{goal}$ = \textbf{BestBranchNode}($\mathbf{V}_{c}$);
    \Else 
    \State $\mathbf{p}_{goal}$ = \text{Select from }$\mathbf{VL}$;
    \EndIf

    \\\textbf{Return} $\mathbf{p}_{goal}$;
    
	\end{algorithmic}

\end{algorithm}


\section{Trajectory Optimization in 3DGS Map}
\label{sec: Trajectory Optimization}
3DGs's explicit representation and precise geometry make safe robot navigation with the 3DGS map possible.
Leveraging the differentiable nature of 3D Gaussian, we devise a safety constraint with the 3DGS map, and integrate it into a quadrotor trajectory optimization framework.

\subsection{Safety Constraint with 3DGS}
\label{subsec:Safety Constraint}
\begin{figure}[b]
    \begin{center}
        \includegraphics[width=1.0\columnwidth]{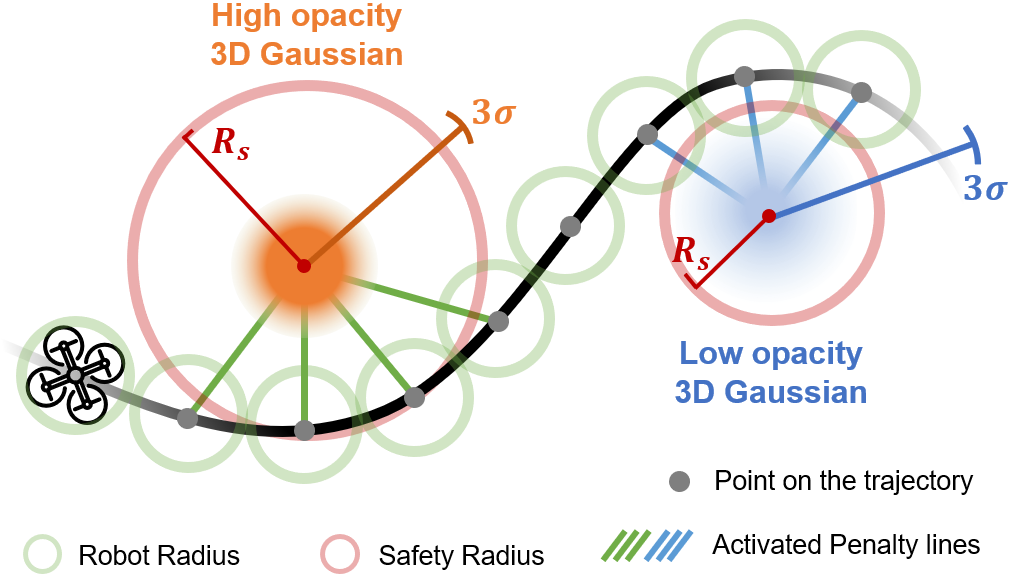}
    \end{center}
    \caption{The collision-avoidance cost applied on the trajectory with different opacity Gaussian. Each point on the trajectory is hoped to be at a distance greater than a safety radius $R_s$ from the mean point of the Gaussian. $R_s$ is weighted by the opacity $o$ of different Gaussians. }
    \label{fig:collision}
    \vspace{0.0cm}
\end{figure}

In 3DGS, Gaussians are defined with opacity as presented in Sec.~\ref{subsec:3DGS Map Representation}.
The opacity measures the probability of light being obstructed while passing through an object.
We assume the probability of terminating a light ray provides a strong indication of the probability of terminating a mass particle. 
Thus, for a robot pose $p$ and a certain Gaussian with opacity $o$, we formulate a chance constraint to ensure safety:
\begin{equation}
\label{equ:chance constraint}
  \alpha(p, o) < c_{thr},
\end{equation}
where $\alpha(\cdot)$ presents the opacity function defined in Eq. \ref{eq:opacity}, and $c_{thr}$ presents the threshold of collision probability. 
$c_{thr}$ is equal to the value of $\alpha(\cdot)$ at a distance of $(3r + R_{robot})$ ($3\sigma$ rule) from its mean $\mu$ when the opacity $o=1$:
\begin{equation}
\label{equ:c_cost}
  c_{thr} = \exp{\left(-\frac{3r + R_{robot}}{2r^2}\right)},
\end{equation}
where $R_{robot}$ is the geometric bounding sphere radius.
Intuitively, it means that we hope every point on the trajectory is at a distance greater than a safety radius $R_s$ from the Gaussian mean point. $R_s$ is weighted by $o$ of the Gaussian, and equals to $(3r + R_{robot})$ when $o=1$.

For the follow-up trajectory optimization, we provide the corresponding collision-avoidance cost for each point $p$ on the trajectory as 
\begin{equation}
\label{equ:c_cost}
  \mathcal J_{c}(p) = \sum_{i=0}^{k}f(\alpha_i(p,o_i)-c_{thr}),
\end{equation}
where $f(x)=\max\left( x,  0\right)^3$, and $k$ is the number of nearby Gaussian elements in the 3DGS map.
The collision-avoidance cost applied on the points on the trajectory during optimization with different opacity Gaussian is shown in Fig.~\ref{fig:collision}.
This differentiable cost is friendly for the follow-up trajectory optimization with analytical gradient written as 
\begin{equation}
\label{equ:c_cost}
  \frac{\partial \mathcal J_{c}(p)}{\partial p} = \sum_{i=0}^{k}3(\alpha_i(p)-c_{thr})^2 o_i exp{\left(-\frac{|p-\mu_i|^2}{2r_i^2}\right)} (\frac{\mu_i-p}{2r_i^2}).
\end{equation}


\subsection{Trajectory Optimization Formulation}
\label{subsec:Trajectory Optimization Formulation}
Aimed to generate full-state collision-free and dynamic-feasible trajectories for quadrotors, we use MINCO \cite{wang2022geometrically} as trajectory representation and optimize spatial-temporal trajectories in a reduced space with differential-flat outputs $\mathbf{z}=[\mathbf{p}^T,\phi]^T \in \mathbb{R}^3 \times \operatorname{SO}(2)$, where $\phi$ is the Euler-yaw angle and position $\mathbf{p}=[p_x,p_y,p_z]^T$.
And we further define the flat outputs and their derivatives $\mathbf z^{[s-1]} \in \mathbb{R}^{ms}$ as
$\mathbf z^{[s-1]} \coloneqq (\mathbf z^T, \dot{\mathbf z}^T,...,{\mathbf z^{(s-1)}}^T)^T$.
To generate a trajectory $\mathbf z(t):[0, T] \mapsto \mathbb{R}^m$, we formulate the trajectory optimization problem as
\begin{subequations}
\label{eq:traj opt problem}
  \begin{align}
    \min_{\textbf{z}, \textbf{T}} & \label{eq:cost function}~\mathcal J_E = \int_{0}^{T} {\norm{\mathbf z^{(s)}(t)}^2} \df{t} + \rho T,\\
    s.t.~          & \label{eq:boundary pos}~\mathbf z^{[s-1]}(0)=\bar{\mathbf{z}}_{s}, ~\mathbf z^{[s-1]}(T)=\bar{\mathbf{z}}_{e},\\
                   & \label{eq:vmax}~\norm{\mathbf{p}^{(1)}(t)} \leq v_{max}, \forall t \in [0,T],\\
                   & \label{eq:amax}~\norm{\mathbf{p}^{(2)}(t)} \leq a_{max}, \forall t \in [0,T],\\
                   & \label{eq:yawmax}~\norm{{\phi}^{(1)}(t)} \leq \phi_{max}, \forall t \in [0,T],\\
                   & \label{eq:safe}~\alpha_i(\mathbf p, o_i)<c_{thr},\forall i\in \{1,\ldots,k\}, \forall t \in [0,T],
  \end{align}
\end{subequations}
where Eq. \ref{eq:cost function} trade off the smoothness and aggressiveness, and $\rho$ is the time regularization parameter. Here we adopt $s=3$ for jerk integral minimization. Eq. \ref{eq:boundary pos} is the boundary conditions at start and end time. $\bar{\mathbf{z}}_{s}$ and $\bar{\mathbf{z}}_{e}$ are the initial and end state, respectively. Eq. \ref{eq:vmax}, Eq. \ref{eq:amax} and Eq. \ref{eq:yawmax} are the dynamic feasibility constraints, where $v_{max}$, $a_{max}$ and $\phi_{max}$ are the velocity, acceleration and yaw rate limits. Eq. \ref{eq:safe} is the safe constraint defined in Eq. \ref{equ:chance constraint}. $\alpha_i(\cdot)$ is the opacity function of the $i$-th Gaussian element with opacity $o_i$.

This problem can be transformed into an unconstrained optimization problem \cite{wang2022geometrically} written as
\begin{equation}
\label{equ:unconstainted problem}
    \begin{aligned}
        \min_{
              \begin{scriptsize}
                \textbf{z},\textbf{T}
              \end{scriptsize}
             }  
          ~\mathcal J_E + \int_{0}^{T}  \mathcal{J}_{\mathcal G} dt,
    \end{aligned}
\end{equation}
where $\mathcal{J}_{\mathcal G}$ is the penalty function corresponding to the inequality constraints Eq. \ref{eq:vmax}, Eq. \ref{eq:amax}, and Eq. \ref{eq:safe}. And $\mathcal{J}_{\mathcal G}$ includes $\mathcal{J}_{c}$ defined in Eq. \ref{equ:c_cost}. With analytical gradients, the problem is then efficiently solved by the L-BFGS \cite{liu1989limited}.

%% file: SEC5_experiments.tex
\begin{figure*}[th]
	\begin{center}
		\includegraphics[width=1.96\columnwidth]{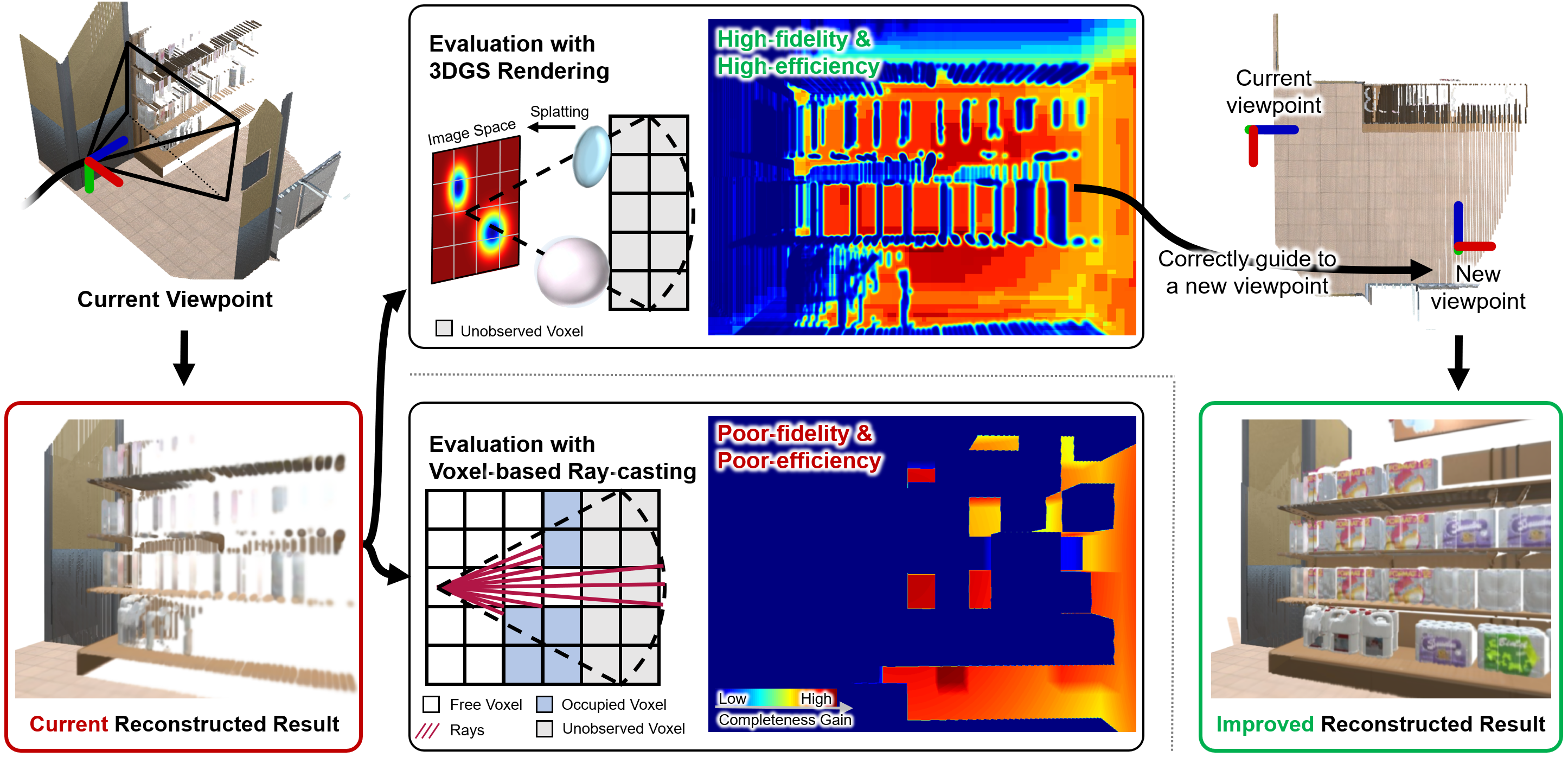}
	\end{center}
	\vspace{-0.2cm}
	\caption{
		\label{fig:abla1} Qualitative comparison of the completeness evaluation between using 3DGS rendering and using voxel-based ray-casting. When the robot arrives at the current viewpoint, due to its oblique view angle, the observation of items on the left front shelf is incomplete. The evaluation with 3DGS rendering is high-fidelity and high-efficiency, while the evaluation with voxel-based ray-casting is coarse and time-consuming. The fine completeness evaluation can correctly guide the robot to collect new information for improvement.  }
	\vspace{-0.0cm}
\end{figure*}

\section{Experiments}
\subsection{Implementation Details}
We run our active reconstruction system on a desktop PC with a 2.90GHz Intel i7-10700 CPU and an NVIDIA RTX 3090 GPU. 
And an additional laptop PC with a 2.50 GHz AMD Ryzen 9 7945HX and an NVIDIA GeForce RTX 4080 Laptop GPU is utilized to execute the high-fidelity simulation developed with Unity. The two devices are connected via a wired network connection. In Unity, the quadrotor equipped with an RGB-D sensor will provide real-time RGB-D images with a resolution of $640 \times 480$ and a perceptual range from $0.5m$ to $3m$. We add a uniform distribution noise of 2cm to the depth and assume the corresponding camera poses of the images are known.

The 3DGS mapping module builds upon SplaTam \cite{keetha2023splatam} by incorporating a real-time data streaming format. 
For view planning, we evaluate 10 viewpoints at each iteration and select the branch with the optimal viewpoint as the seed for the next iteration. 
For trajectory optimization, the robot radius is fixed at 0.5m. And the safety constraint is computed by considering the 3DGS near the initial trajectory within the duration of $[0s,1s]$, selected using the Axis-aligned Bounding Box (AABB) method. The maximum velocity limit is $1.0 m/s$, the maximum acceleration limit is $2.0 m/s^2$, and the maximum yaw rate limit is $\pi rad/s$.

\subsection{Simulation Result and Analysis}
To validate our proposed method, we build a high-fidelity simulation environment via Unity engine. As shown in Fig.~\ref{fig:simu_res}, this $22.0m\times 14.0m \times 3.2m$ supermarket scene contains a variety of items with rich texture information.
We present the whole reconstruction process and the trajectory of the quadrotor. 
The quadrotor takes 343 seconds to complete the whole reconstruction.
The reconstructed details are also demonstrated through rendered RGB and depth images.
We can see from the reconstruction results that the reconstruction of the entire scene is complete and high-fidelity, retaining rich texture and structural information, and exhibiting a strong sense of realism.

\subsection{Comparision and Ablation Study}
To validate the effectiveness of the proposed reconstruction evaluations, we compare our method with traditional ones and conduct an ablation study.

\subsubsection{Completeness Evaluation}
Given a viewpoint, traditional methods for computing information gain typically rely on voxel-based raycasting \cite{isler2016information, dang2019graph, zhou2021fuel}. This involves maintaining a grid map that represents observed and unobserved areas, and performing raycasting at candidate viewpoints to measure the volume of unobserved areas. However, this method is limited by the voxel resolution for occupied and unobserved region representation, and its computational complexity is affected by discrete sampling steps. 
In contrast, we integrate the completeness evaluation calculation into the splatting process. Leveraging efficient Gaussian sorting and precise description of occupied geometry, we achieve high-fidelity high-efficiency completeness gain calculation.
Fig.~\ref{fig:abla1} shows an instance of the completeness gain calculation by different methods. 
Tab.~\ref{tab:benchmark} compares the computation speeds under various voxel resolutions, highlighting the notably higher efficiency of our 3DGS-based method. In the experiments, the raycasting step for the voxel-based method is half of the voxel resolution.

\begin{table}[t]
\centering
\caption{Completeness Evaluation Methods Comparison}
\label{tab:benchmark}
\renewcommand\arraystretch{1.5}
\setlength{\tabcolsep}{2.7mm}{
\begin{tabular}{|c|c|c|c|}
\hline
    \multirow{2}{*}{\makecell[c]{Voxel\\Resolution (m)}}  & \multirow{2}{*}{Scenario} & \multicolumn{2}{c|}{Time (ms)} \\ \cline{3-4}
            &          & Voxel-based Raycast & \textbf{Ours} \\ \hline
    \multirow{2}{*}{0.1}  & Sparse   & 347.32      & 1.86          \\ \cline{2-4} 
                          & Dense    & 342.10      & 2.11           \\ \cline{1-4}
    \multirow{2}{*}{0.15}  & Sparse   & 226.29     & 1.83          \\ \cline{2-4} 
                          & Dense    & 230.21      & 2.33         \\ \cline{1-4}
    \multirow{2}{*}{0.2}  & Sparse   & 183.36      & 1.71          \\ \cline{2-4} 
                          & Dense    & 176.01      & 2.31            \\ \cline{1-4}
\end{tabular}}
\vspace{-1.2cm}
\end{table}

\begin{figure*}[th]
	\begin{center}
		\includegraphics[width=1.98\columnwidth]{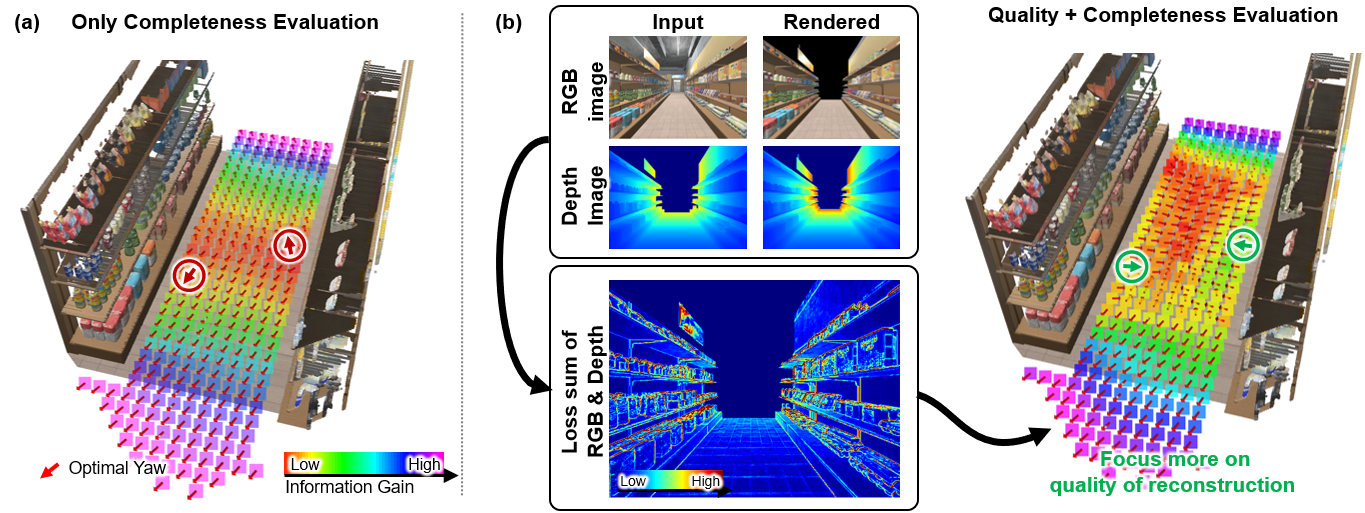}
	\end{center}
	\vspace{-0.0cm}
	\caption{
		\label{fig:abla_2} Ablation of the quality gain. (a). The information gain regarding only completeness at the height of $z=1~m$. Optimal yaw angles corresponding to the candidate viewpoints point towards unobserved areas. (b). Considering both quality and completeness in the information gain. It can be observed that, for viewpoints around two shelves, the quality gain tends to encourage further observation of shelves that can still improve the reconstruction quality. }
	\vspace{-0.5cm}
\end{figure*}

\subsubsection{Quality Evaluation}
To validate the impact of quality gain, we designed ablation experiments to calculate the information gain of candidate viewpoints with and without quality gain.
And we further compute their corresponding optimal yaw angles.
As the result shown in Fig.~\ref{fig:abla_2}, the quality gain correctly guides the generation of the information gain and the optimal yaw angle.
With quality consideration, our active reconstruction system can improve the regions of the built scene with poor geometry and texture.
